\documentclass[10pt,twocolumn,letterpaper]{article}

\usepackage{cvpr}
\usepackage{times}
\usepackage{subcaption}
\usepackage{graphicx}
\usepackage{float}
\usepackage{amsmath}
\usepackage{amsfonts}
\usepackage{amssymb}
\usepackage{textcomp}
\usepackage{siunitx}

\usepackage[pagebackref=true,breaklinks=true,letterpaper=true,colorlinks,bookmarks=false]{hyperref}

\cvprfinalcopy 

\def\cvprPaperID{4829} 

\ifcvprfinal\fi

\newcommand{\DMG}[1]{{\textcolor{black}{{#1}}}} 
\newcommand{\CSe}[1]{{\textcolor{black}{{#1}}}} 
\begin{document}
\title{Privacy Protection in Street-View Panoramas using Depth \\ and Multi-View Imagery} %
\author{Ries Uittenbogaard$^{1}$, \hspace{2mm} $^{*}$Clint Sebastian$^{2}$, \hspace{2mm} Julien Vijverberg$^{3}$, \\
Bas Boom$^{3}$, \hspace{2mm} Dariu M. Gavrila$^{1}$, \hspace{2mm} Peter H.N. de With$^{2}$ \\ \\
$^{1}$Intelligent Vehicles Group, TU Delft, \hspace{3mm}$^{2}${VCA Group, TU Eindhoven}, \hspace{2mm}$^{3}$Cyclomedia B.V\\
{\tt\small c.sebastian@tue.nl} \\
$^{*}$\small{corresponding author}
}

\maketitle

\begin{abstract}

The current paradigm in privacy protection in street-view images is to detect and blur sensitive information. In this \DMG{paper}, we propose a framework that is an alternative to blurring, which automatically removes and inpaints moving objects (\DMG{e.g.} pedestrians, vehicles) in street-view imagery. We propose a novel moving object segmentation algorithm exploiting consistencies in depth across multiple street-view images that are later combined with the results of a segmentation network. The detected moving objects are removed and inpainted with information from other views, to obtain a realistic output image such that the moving object is not visible anymore. We evaluate our results on a dataset of 1000 images to obtain a peak noise-to-signal ratio (PSNR) and $L_1$ loss of \SI{27.2}{\decibel} and 2.5\%, respectively. 
\DMG{To assess overall quality, we also report the results of a survey conducted on 35 professionals, asked to visually inspect the images whether object removal and inpainting had taken place}. The inpainting dataset will be made publicly available \DMG{for scientific benchmarking purposes at \url{https://research.cyclomedia.com/}.}
 
\end{abstract}
\vspace{-5mm}
\section{Introduction}
\label{sec:Introduction}
In recent years, street-view services such as Google Street View, Bing Maps Streetside, Mapillary
have systematically collected and hosted millions of images. Although these services are useful, they have been withdrawn or not updated in certain countries~\cite{gsvstopped, gsvindia},  due to serious privacy concerns. The conventional way of enforcing  privacy in  street-view  images is  by  blurring  sensitive information such as faces and license plates. However, this has several drawbacks. First, the blurring of an object like a face might not ensure that the privacy of the person is sufficiently protected. 
\begin{figure}[!t]
    \centering
    \begin{subfigure}{0.24\textwidth}
        \centering
        \includegraphics[width=118pt, height=106pt, trim= {0 5pt 0 5pt}, clip]{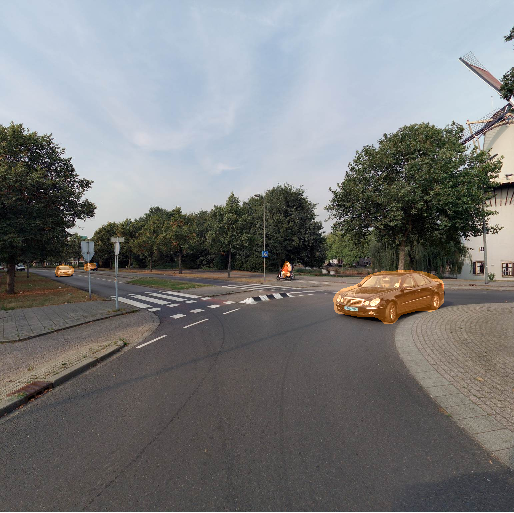}
    \end{subfigure}%
    \begin{subfigure}{0.24\textwidth}
        \centering
        \includegraphics[width=118pt, height=106pt, trim= {0 5pt 0 5pt}, clip]{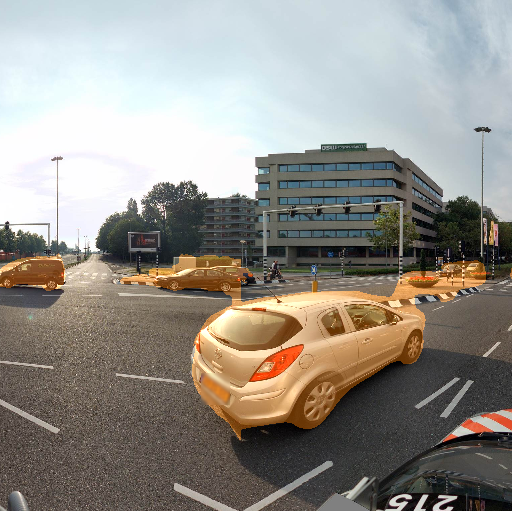}
    \end{subfigure}
    \begin{subfigure}{0.24\textwidth}
        \centering
        \includegraphics[width=118pt, height=106pt, trim= {0 5pt 0 5pt}, clip]{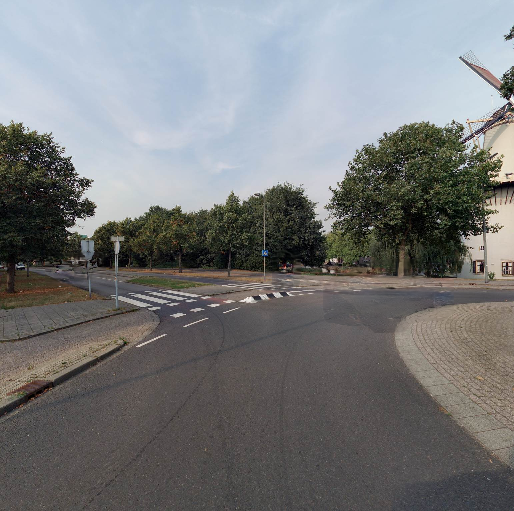}
    \end{subfigure}%
    \begin{subfigure}{0.24\textwidth}
        \centering
        \includegraphics[width=118pt, height=106pt, trim= {0 5pt 0 5pt}, clip]{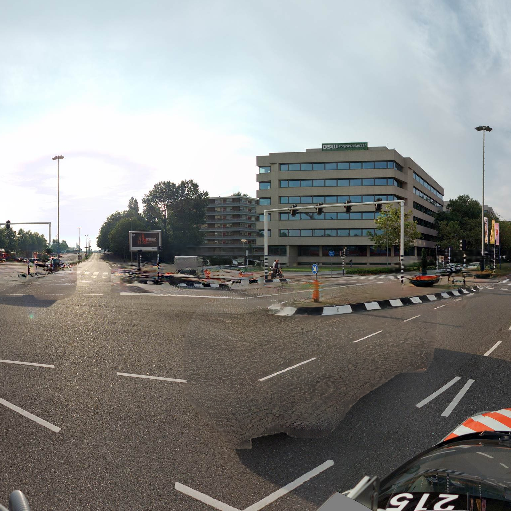}
    \end{subfigure}
    \caption{Example of moving object segmentation (top) and the results after inpainting (bottom). The regions that are highlighted in orange are removed and inpainted.}
    \label{fig:Intro}
\end{figure}
\setlength{\textfloatsep}{10pt plus 1.0pt minus 5.0pt}
The clothing, body structure, location and several other aspects \DMG{can lead} to the identity of the person, even if the face is hidden. Second, blurring objects creates artifacts that is undesirable in application when consistent view of the infrastructure is required. For commercial purposes such as change detection and localization of objects from street-view imagery, blurring limits the scope of these applications.

\begin{table*} [t!] 
\begin{center}
  \begin{tabular}{ | c | c | c | c | c | c | }
    \hline
    \textbf{Method} & \textbf{Input data} & \textbf{\shortstack{Image matching}} & \textbf{Detection} & \textbf{Class/No. of objects} & \textbf{Inpainting method} \\
    \hline
    Flores \textit{et al.}  & \shortstack{Grayscale, \\ Multi-view } & \shortstack{Homography \\ computation using \\ SIFT \& RANSAC} & \shortstack{Leibe's detector \\ (no moving object \\ detection, manual)} & \shortstack{One pedestrian \\ per image }  & \shortstack{Homography-based \\ warping + Davis \\ compositing } \\  
    \hline
    Ours & \shortstack{RGB-D, \\ Multi-view} & \shortstack{Real world positions \\ from GPS, IMU \\ with camera intrinsics} & \shortstack{Novel deep learning \\ based moving object \\ segmentation } & \shortstack{Any number of \\ objects and \\ classes per image} & \shortstack{Reprojection from \\ multiple views + \\ multiview inpainting  \\ GAN} \\
    \hline
  \end{tabular}
  \vspace{-1mm}
  \caption{Comparison of the proposed vs. closest related method.} 
  \label{comparison}
\end{center}
\vspace{-8mm}
\end{table*}

\DMG{It is therefore} desirable to use a method that completely removes the identity-related information. In this paper, we present a method that automatically segments and replaces moving objects from sequences of panoramic street-view images by inserting a realistic background. Moving objects are of primary interest because authentic information about the background is present in the other views (we assume that most moving objects are either vehicles or pedestrians). It is risky to inpaint static objects after removing them, as they may remove important or introduce new information from context. We do not focus on inpainting static objects such as parked cars, standing pedestrians etc. as no authentic information of the background can be derived. However, inpainting from spatial context is viable solution for non-commercial applications. Using a segmentation network to detect a moving object is a challenging task, since it needs to learn to distinguish moving from static objects. To simplify this problem, we generate a prior segmentation mask, exploiting the consistencies in depth. The proposed moving object detection algorithm, combines with results from a standard segmentation algorithm to obtain segmentation masks for moving objects. Finally, to achieve an authentic completion of the removed moving objects, we use inpainting Generative Adversarial Network (GAN) that utilizes multi-view information. While multi-view GAN has been explored for synthesizing an image to another view~\cite{chen2017multi, Sun2018MultiViewTV}, to the best of our knowledge, this is the first work that exploits multi-view information for an inpainting GAN network.

\section{Related Work}
\label{sec:Related Work}
\paragraph{Privacy protection} Several approaches have been proposed for privacy protection in street-view imagery~\cite{frome2009large, flores2010removing, agrawal2011person, nodari2012digital, padilla2015visual}. The most common way to hide sensitive information is to detect the objects of interest and blur them~\cite{sebastianprivacy2019}. However, few works have explored the removal of privacy-sensitive information from street-view imagery for privacy protection. Flores and Belongie~\cite{flores2010removing} detects and inpaints pedestrians from another view~(details in Table~\ref{comparison}).
Similarly, Nodari~\etal~\cite{nodari2012digital} also focuses on pedestrians removal. However, they remove the pedestrian with a coarse inpainting of the background. This is followed by replacement of the inpainted region with a pedestrian obtained from a controlled and authorized dataset. Although this method ensures the privacy, the replaced pedestrians tend to appear unrealistic with respect to the background.
\vspace{-5mm}
\paragraph{Object detection} Due to the progress in deep learning, there have been significant improvements
in object detection~\cite{ren2015faster, dai2016r, liu2016ssd}. Detecting objects of interest provides a reliable way to localize faces and license plates. Similarly, for precise localization, semantic segmentation offers a better alternative to bounding boxes~\cite{long2015fully, badrinarayanan2015segnet2, chen2018deeplab}. Hence, we rely on semantic segmentation approaching pixel accuracy, as it requires fewer pixels to be replaced during inpainting. We obtain our segmentation masks through a combination of the proposed moving object segmentation and segmentation from a fully convolutional deep neural network. 

In the recent years, LiDAR systems has become ubiquitous for applications like self-driving cars and 3D scene reconstruction.  Several moving object detection methods rely on LiDAR as it provides rich information about the surroundings~\cite{Ferri2015DynamicOD, Azim2012DetectionCA, Cho2014AMF, Yan2014AutomaticEO}. Few approaches convert LiDAR-based point-clouds into 3D occupancy grids or voxelize them~\cite{Ferri2015DynamicOD, Azim2012DetectionCA}. These are later segregated into occupied and non-occupied building blocks. The occupied building blocks are grouped into objects and are tracked over time to determine moving objects~\cite{takabe2016moving}. Fusion of both LiDAR and camera data has also been applied for object detection~\cite{Yan2014AutomaticEO, Cho2014AMF, takabe2016moving}. In this case, consistency across both image and depth data (or other modalities) in several frames are checked to distinguish static and moving objects. 
\vspace{-5mm}
\paragraph{Inpainting} Prior works have tried to produce a realistic inpainting by propagating known structures at the boundary into the region that is to be inpainted~\cite{barnes2009patchmatch}. However, in street-view imagery, this is a challenging task especially when it has large holes and require complex inpainting. Therefore, few results relied on an exemplar or multi-view based methods~\cite{criminisi2004region, buyssens2015exemplar}.  State-of-the-art of inpainting methods adopt Generative Adversarial Networks (GANs) to produce high-quality inpainted images~\cite{goodfellow2014generative}. GANs are often applied for problems such as image inpainting~\cite{Pathak2016ContextEF, Iizuka2017GloballyAL, Yu2018GenerativeII}, image-to-image translation~\cite{zhu2017unpaired, wang2017highres}, conditional image translation~\cite{isola2016image, sebastian2018conditional} and super-resolution~\cite{johnson2016perceptual, ledig2016photo}. 

\begin{figure*}[t]
    \centering
    \includegraphics[width=500pt, height=200pt]{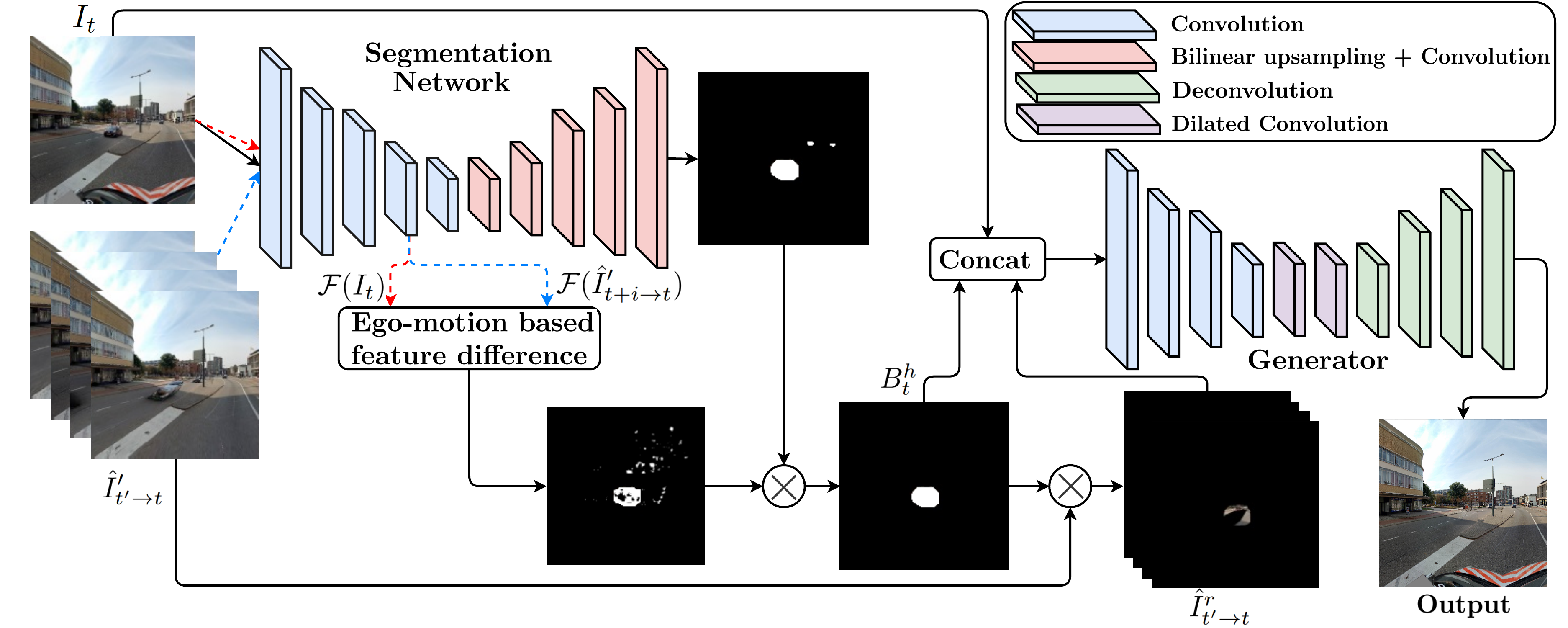}
    \vspace{-7mm}
    \caption{Overview of the proposed method to segment moving objects and inpaint them from other views. The input image is first fed to segmentation network to produce the segmentation mask of both moving and static objects. The difference of the convolution features of the reprojected images and input image is used to find the moving objects in the segmentation mask. The original input image ($I_t$), moving object segmentation mask ($ B_t^h$) and the reprojected images with regions active in the segmentation mask ($\hat{I}^r_{t' \rightarrow t}$) is fed to the generator. Note that the discriminators networks are not shown for simplicity.}
    \label{fig:Overview}
    \vspace{-2ex}%
\end{figure*}

Different approaches have been proposed for inpainting images using deep neural networks. Pathak~\etal proposed one of the first methods that utilized a deep neural network~\cite{Pathak2016ContextEF}. They applied a combination of both reconstruction and adversarial losses to improve the quality of the inpainted images. This was improved in ~\cite{Iizuka2017GloballyAL}, using dilated convolutions and an additional local discriminator. To improve the quality of details in the output image, Zhao~\etal proposes to use a cascade of deep neural networks~\cite{Zhao2017ADC}. The network first inpaints with a coarse result, followed by a deblurring-denoising network to refine the output. A multi-stage approach for inpainting is also proposed in~\cite{Yu2018GenerativeII}.  Yu~\etal introduces an attention based inpainting layer that learns to inpaint by copying features from context. They also introduce a spatially discounted loss function in conjunction with improved Wasserstein GAN objective function~\cite{improvedwgan2017} to improve the inpainting quality. 

Although inpainting based on context produces plausible outputs for accurate image completion, GANs may introduce information that is not present in reality. This is undesirable, especially in commercial applications, where objects of interest are present or accurate localization are required. A reasonable idea here is to utilize information from other views as a prior. Other views provide a better alternative than inpainted information from scratch. An example would be the case when an object of interest (\DMG{e.g.} traffic-sign, bill-board) is occluded by a car or person. After moving object detection, using a GAN to inpaint the hole from context information would remove the object of interest. However, multi-view information could alleviate this problem as the object of interest is visible in the other views. Our \DMG{main paper contributions are}:
\begin{itemize}
    \item We propose a new \DMG{multi-view framework for detecting and inpainting moving objects, as an alternative to blurring in street-view imagery}  
    \item We introduce a new moving object detection algorithm based on convolutional features that exploits depth consistencies from a set of consecutive images.
    \item We train an inpainting GAN that utilizes multi-view information to obtain authentic and plausible results.
\end{itemize}

\section{Method}
\label{sec:Method}
First, we construct a method that combines standard segmentation with a novel moving object segmentation, which segments the moving objects from a consecutive set of images that have a large baseline. The moving object is estimated using an ego-motion based difference of convolutional features.
Second, we use a multi-view inpainting GAN to fill-in the regions that are removed from moving object detection algorithm. The overview of the proposed framework is shown in Fig~\ref{fig:Overview}.

\subsection{Moving Object Segmentation}
\label{sec:Method Moving}
\vspace{-1mm}
For supervised segmentation, we apply a Fully Convolutional VGGNet (FC-VGGNet), due to its simplicity and the rich features that are used for moving object segmentation. We make slight modifications to VGGNet-16 by removing fully-connected layers and appending bilinear upsampling followed by convolution layer in the decoder. To create the segmentation mask for a specific image $I_t$ at time $t$, the detection algorithm also uses the two images captured before and after it, \ie the RGB images at time $t-2, \ldots, t+2$.
Finally, from the LiDAR-based point cloud and the positions of each recording, the depth images for these time steps are created. Note that the RGB and the depth image are not captured at the same time. Hence, the moving objects are not at the same positions.
Reprojecting the image $I_{t'}$ to the position of image $I_t$ is achieved using its respective depth images $D_{t'}$ and $D_t$. Employing the depth images in conjunction with recorded GPS positions leads to real-world pixel positions $\vec{p}'_{t'}$, $\vec{p}_t$, resulting in the defined image reprojection $\hat{I}_{t' \rightarrow t}$.
\begin{figure}[!t]
    \centering
    \begin{subfigure}{0.25\textwidth}
        \centering
        \includegraphics[width=123pt, height=95pt]{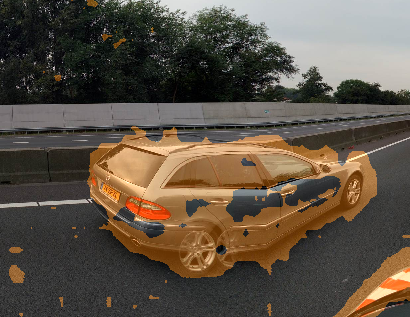}
    \end{subfigure}%
    \begin{subfigure}{0.25\textwidth}
        \centering
        \includegraphics[width=123pt, height=95pt]{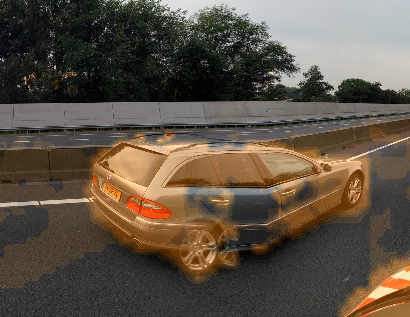}
    \end{subfigure}
    \caption{Results from features extracted from FC-VGGNet (first) and VGGNet (second). Features from FC-VGGNet are well-localized and have strong activations.}
    \vspace{-2mm}
    \label{fig:vgg_vs_segnet}
\end{figure}
Evidently, some pixels in $\hat{I}_{t' \rightarrow t}$ cannot be filled due to occlusions.
These pixels are replaced by the pixel values of $I_t$ by comparing the distance between the real-world points to an heuristically defined threshold $\epsilon$. This reprojection with a threshold is given by
\begin{equation}
    \hat{I}'_{t' \rightarrow t} = 
    \left\{ 
    \begin{array}{ll}
         \hat{I}_{t' \rightarrow t} & \mbox{if } ||\vec{p}_t - \vec{p}'_{t'}|| < \epsilon, \\
         I_t & \mbox{otherwise.}  
    \end{array} 
    \right.
    \vspace{-1mm}
\end{equation}
\begin{figure*}[t!]
    \begin{subfigure}[b]{1\textwidth}
        \centering
        \includegraphics[width=\textwidth, trim= {0 10pt 0 10pt}, clip]{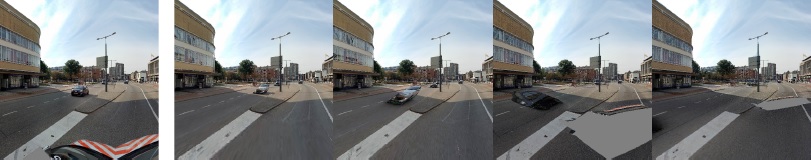}
    \end{subfigure}
    \begin{subfigure}[b]{1\textwidth}
        \centering
        \includegraphics[width=\textwidth, trim= {0 10pt 0 10pt}, clip]{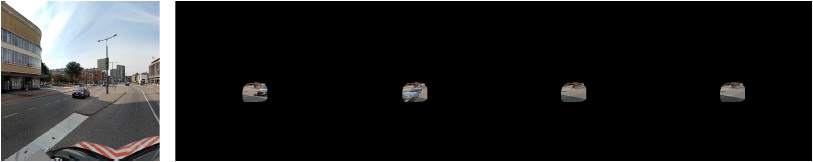}
    \end{subfigure}
    \begin{subfigure}[b]{1\textwidth}
        \centering
        \includegraphics[width=\textwidth, trim= {0 10pt 0 10pt}, clip]{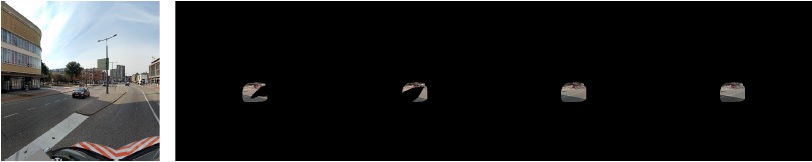}
    \end{subfigure}
    \caption{Images in the first column are the input images ($I_t$) at time $t$. The first row contains images $\hat{I}'_{t' \rightarrow t}$ for $t' \in \{ -2, -1, +1, +2\}$ that is projected to the view point of $I_t$. Results in the second row are obtained after removal of regions around the area of interest. Finally, the third row ($\hat{I}^r_{t' \rightarrow t}$) is obtained after removal of moving objects from other views. }
    \label{fig:ReprojectionMasking}
    \vspace{-2ex}%
\end{figure*}
%
Fig.~\ref{fig:ReprojectionMasking} (Row 1) shows an example of reprojection for 4~neighbouring recordings of 5~consecutive images. A simple pixel-wise comparison between $\hat{I}'_{t' \rightarrow t}$ and $I_t$ yields poor segmentation results, due to slight variations in the position of the car. We have empirically found that patch-based comparison utilizing pretrained network features produce better results than conventional pixel-wise and patch-based features. Feature extraction is often applied to generate descriptors or rich representations for applications such as image retrieval and person reidentification~\cite{Arandjelovic2018NetVLAD, noh2017largescale,liu2018part}. However, here we utilize the extracted features to obtain the moving objects. Instead of using VGG~\cite{Simonyan2015VGG} or other pretrained network features, we extract features from FC-VGGNet that is trained to detect static objects, as it is easier to reuse the same network for moving object segmentation. Besides the simplicity of relying on a single network and higher performance, this also speeds up the pipeline.  High-dimensional features $\mathcal{F}(I) \in \mathbb{R}^{64 \times 256 \times 512}$ are extracted from the output of the $4^{th}$ convolution block.
The moving object segmentation score is the average of the $L_1$ norms between each of the projected images and $I_t$, which is specified by
\vspace{-3mm}
\begin{equation}
    s_t^{1/8} = \frac{1}{4} \sum_{i \in \{ -2, -1, 1, 2\}} || \mathcal{F}(I_t) - \mathcal{F}(\hat{I}'_{t+i \rightarrow t}) ||_1 \ ,
    \vspace{-3mm}
\end{equation}

where $s_t^{1/8}$ is upsampled by factor of 8 to obtain a scoring mask $s_t$ of the original input size of 512~$\times$~2048. Examples of the outputs $s_t$ is shown in Fig.~\ref{fig:vgg_vs_segnet}. To generate accurate segmentation masks of moving objects, FC-VGGNet is trained on the 4~classes that includes both moving and static objects. For each of the extracted objects from the final output segmentation masks $m_t$ of FC-VGGNet, we compute the element-wise product with the scoring mask $s_t$. We classify an object as moving if the mean of the scores exceeds a threshold $\tau$ in the given object area $\mathcal{A}$, yielding
\vspace{-2mm}
\begin{equation}
    \frac{1}{n}\sum_{(x,y) \in \mathcal{A}} s_t(x,y) \cdot m_t(x,y)  > \tau,
    \label{eq:IsMoving}
    \vspace{-2mm}
\end{equation}
where $n$ is the number of elements in $\mathcal{A}$.
The value of the threshold $\tau$ is discussed in Section~\ref{sec:Experiments Moving-Object Segmentation}. 
\subsection{Inpainting}
\label{sec:MethodInpainting}
After obtaining the segmentation masks from the moving object segmentation, we remove the detected objects. With respect to previous approaches, our method requires inpainting from other views that serve as a prior. Our input images are also larger (512~$\times$~512 pixels) compared to ~\cite{Yu2018GenerativeII} (256~$\times$~256) and hence we added an additional strided convolution layer in the generator and two discriminators. Our inputs consist of an RGB image with holes $I_t^h$, the binary mask with the holes $B_t^h$ obtained from the moving object detection and RGB images $\hat{I}^r_{t' \rightarrow t}$ that are projected from the other views. The images $\hat{I}^r_{t' \rightarrow t}$ are obtained from reprojection after removal of moving object from other views ($I^r$ denoting removed objects) in the regions where holes are present in the binary mask $B_t^h$. This is shown in the third row of the Fig.~\ref{fig:ReprojectionMasking}. The final input to the generator is a 16-channel input. 

 The 16-channel input is fed to the coarse network from~\cite{Yu2018GenerativeII} to produce the final output. We follow a similar approach, however, the refinement network is not used as no performance improvement is observed. This occurs as the input contains sufficient prior information which alleviates the need to produce a coarse output. We also observe that we need to train for longer period of time with a single-stage network to reach the performance of the two-stage network.  We follow the same strategy in~\cite{sebastian2018conditional, Yu2018GenerativeII} of training multiple discriminators to ensure both local and global consistency. Hence, the output from the network is fed to a global and local discriminator. For training, we use the improved WGAN objective~\cite{improvedwgan2017} along with a spatially discounted reconstruction loss~\cite{Yu2018GenerativeII}. The final training objective $\mathcal{L}$ with a generator $\textit{G}$ and discriminator networks $\textit{D}^c$ (where $\textit{c}$ denotes the context, global or local discriminator) is expressed :
\vspace{-2mm}
\begin{equation}
    \mathcal{L} = 
    \underset{\textit{G}}{\textnormal{min }} 
    \underset{\textit{D}^{c}}{\textnormal{max }} 
    \mathcal{L}_{\textnormal{WGAN-GP}}^h (\textit{G}, \textit{D}^{c}) 
    + 
    \mathcal{L}_{L1}^d (\textit{G}, \textit{I$_{t}$}),
	\vspace{-1mm}
\end{equation}
where $\mathcal{L}_{\textnormal{WGAN-GP}}^h$ is the WGAN adversarial loss with gradient penalty applied to pixels within the holes and $\mathcal{L}_{L1}^d$ is the spatially discounted reconstruction loss. We follow the same WGAN adversarial loss with gradient penalty in~\cite{Yu2018GenerativeII} for our problem,  
\begin{equation}
	\begin{aligned}
		\mathcal{L}_{\textnormal{WGAN-GP}}^h \big(\textit{G}, \textit{D}\big)  = 
		\mathbb{E}_{\mathbf{\Tilde{x}} \sim \mathbb{P}_{f}}
		[\textit{D}(\mathbf{\Tilde{x}})] 
		-
		\mathbb{E}_{\mathbf{x} \sim \mathbb{P}_{r}}
		[\textit{D}(\mathbf{x})] 
		\\ +  \ \ 
		\lambda\ 
		\mathbb{E}_{\mathbf{\hat{x}} \sim \mathbb{P}_{\mathbf{\hat{x}}}}
        \big[({\left \| {\nabla}_{\mathbf{\hat{x}}} \textit{D}(\mathbf{\hat{x}}) 
        \odot (\mathbf{1 - m})\right \|}_2 - 1)^2 \big],
	\end{aligned}
\end{equation}
where ${\nabla}_{\mathbf{\hat{x}}} \textit{D}(\mathbf{\hat{x}})$ denotes the gradient of $\textit{D}(\mathbf{\hat{x}})$ with respect to $\mathbf{\hat{x}}$ and $\odot$ denotes the element-wise product. The samples $\mathbf{x}$ and $\mathbf{\Tilde{x}}$ are sampled from real and generated distributions $\mathbb{P}_r$ and $\mathbb{P}_f$. $\mathbb{P}_f$ is implicitly defined by $\mathbf{\Tilde{x}} = \textit{G}$([$I_t, B_t^h, \hat{I}^r_{t' \rightarrow t}$]) where [,] denotes the concatenation operation and $t' \in \{ -2, -1, 1, 2\}$ . The sample $\mathbf{\hat{x}}$ is an interpolated point obtained from a pair of real and generated samples. The gradient penalty is computed only for pixels inside the holes, hence, a mask $\mathbf{1 - m}$ is multiplied with the input where the values are 0 for missing pixels and 1 otherwise. The spatial discounted reconstruction loss~\cite{Yu2018GenerativeII} $\mathcal{L}_{L1}^d$ is simply weighted $\mathbf{L1}$ distance using a mask $\mathbf{M}$ and is given as
\vspace{-1mm}
\begin{equation}
	\begin{aligned}
    \mathcal{L}_{L1}^d (\textit{G}, \textit{I$_{t}$}) = 
    \| \mathbf{M} 
    \odot 
    \textit{G}([I_t, B_t^h, \hat{I}^r_{t' \rightarrow t}]) - \mathbf{M} \odot  \textit{I}_t\|_1,
	\end{aligned}
	\vspace{-1mm}
\end{equation}
where each value in the mask $\mathbf{M}$ is computed as ${\gamma}^l$ ($l$ is the distance of the pixel to nearest known pixel). We set the gradient penalty coefficient $\lambda$ and the value $\gamma$ to 10 and 0.99 respectively as in \cite{improvedwgan2017, Yu2018GenerativeII}. Intuitively, the $\mathcal{L}_{\textnormal{WGAN-GP}}^h$ updates the generator weights to learn plausible outputs whereas $\mathcal{L}_{L1}^d$ tries to the reconstruct the ground truth. 

\section{Experiments}
\label{sec:Experiments}
We evaluate our method on the datasets described in the next section. The final results are evaluated using peak signal-to-noise ratio (PSNR), $\mathit{L_1}$ loss and an image quality assessment survey.

\subsection{Datasets}
\label{sec:MethodDataset}
The datasets consists of several high-resolution panoramas and depth maps derived from LiDAR point clouds. Each of the high-resolution panoramas is obtained from a five-camera system that has its focal point on a single line parallel to the driving direction. The cameras are configured such that the camera centers are on the same location, in order to be able to construct a 360\textdegree~panorama. The parallax-free 360\textdegree~panoramic images are taken at every 5-meters and have a resolution of 100-megapixels. The images are well calibrated using multiple sensors (GPS, IMU) and have a relative positioning error less than 2 centimeters between consecutive images. The LiDAR scanner is a Velodyne HDL-32E with 32 planes, which is tilted backwards to maximize the density of the measurement. \CSe{The RGB and LiDAR are recorded together and they are matched using pose graph optimization from several constraints such as IMU, GPS. The point cloud from the LiDAR is meshed and projected to a plane to obtain a depth map. }

The \textbf{segmentation dataset} consists of 4,000 images of 512 $\times$ 512 pixels, 360\textdegree~panoramas along with their depth maps. The dataset is divided into 70\% for training and 30\% for testing.  Our internal dataset consists of 96~classes of objects out of which 22~are selected for training. The 22~classes are broadly segregated into 4~classes as recording vehicle, pedestrians, two-wheelers and motorized vehicles. The \textbf{inpainting dataset} contains of 8,000 images where 1,000 are used for testing. The holes for inpainting have varying sizes (128~$\times$~128 to 384~$\times$~384 pixels) that are placed randomly at different parts of the image. The inpainting dataset will be made publicly available. 

\subsection{Moving-Object Segmentation}

\label{sec:Experiments Moving-Object Segmentation}
We first train FC-VGGNet on our internal dataset across the pre-mentioned 4 sub-classes described in Section~\ref{sec:MethodDataset}. Due to high class imbalances (recording  vehicle (5.3\%),  pedestrians (0.05\%), two-wheelers (0.09\%) and motorized vehicle (3.2\%), other objects (91.4\%)), the losses are re-weighted inversely proportional to the percent of pixels of each class. We observe the best performance at 160~epochs to obtain a mean IoU of 0.583.

\begin{figure}[!t]
    \centering
    \includegraphics[width=0.90\linewidth]{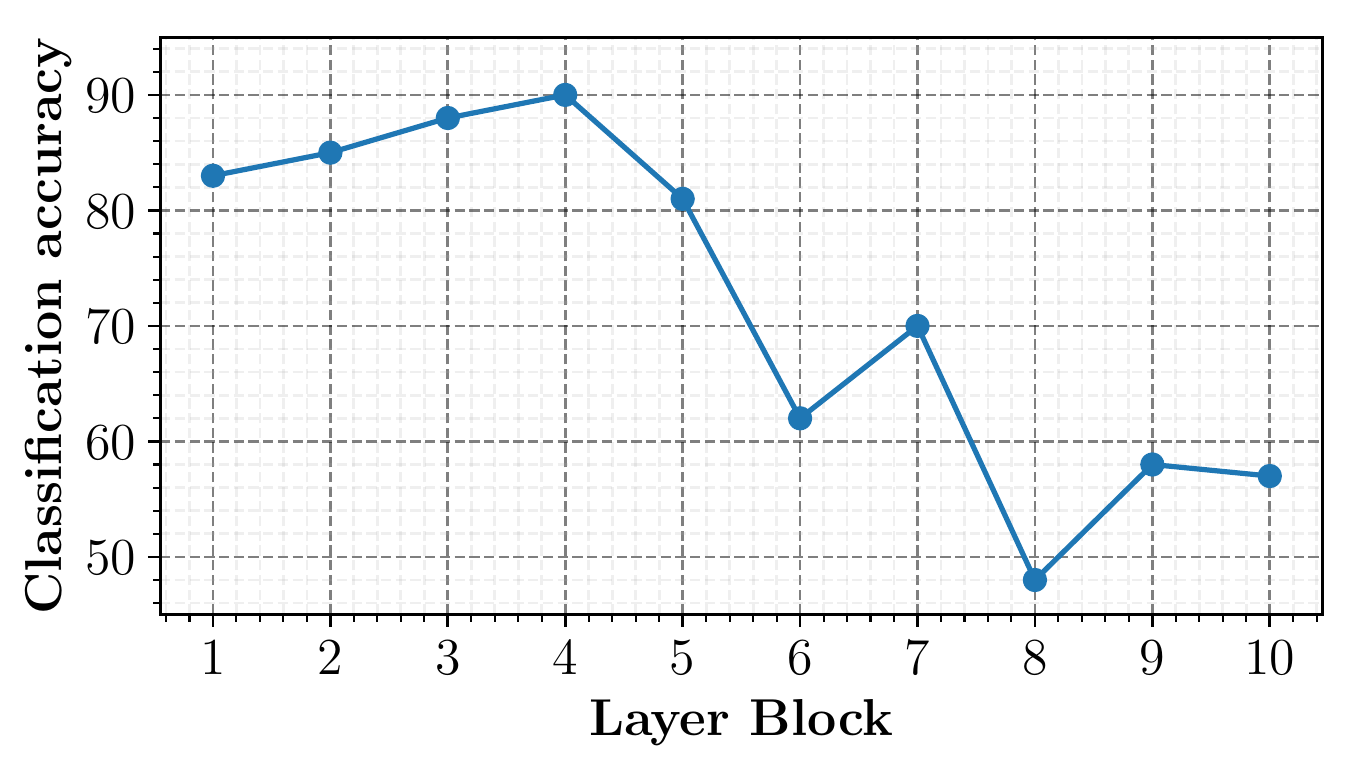}
    \caption{Classification accuracy of extracted objects from moving object segmentation results as moving/non-moving from different layer blocks of FC-VGGNet.}
    \label{fig:SegNetLayerCorrectlyClassified}
\end{figure}

Since large public datasets of moving objects with ground truth are not available, we resort to manual evaluation of the segmentation results. To evaluate the performance of the moving object detection, we select every moving object in 30 random images. We measure the classification accuracy of these extracted moving objects across different layer blocks of FC-VGGNet to determine the best performing layer block. From Fig.~\ref{fig:SegNetLayerCorrectlyClassified}, we can conclude that extracting features $\mathcal{F}(I_t)$ from convolution layers of the decoder (Layers 6-10) of FC-VGGNet leads to worse classification results than using the convolution layers of the encoder (Layers 1-5). The best results are obtained when the outputs are extracted from the fourth convolution layer. We also conducted experiments with a VGG-16, pretrained on ImageNet dataset. We have found that the best results are extracted from the eighth convolution layer, which produces an output of size 28 $\times$ 28 $\times$ 512. However, on qualitative analysis, we have found that the features from the encoder layers of FC-VGGNet offer much better performance than VGG features. This is visualized in Fig.~\ref{fig:vgg_vs_segnet}. The activations are stronger (higher intensities) for moving objects and have lower false positives. This is expected as FC-VGGNet is trained on the same data source as used for testing, whereas VGG is trained on ImageNet. 

It is interesting to observe that features from the shallower layers (earlier layers) of FC-VGGNet perform much better than deeper layers (later layers) for moving object detection.  This is due to features adapting to the final segmentation mask as the network grows deeper. As we compute the $L_1$ loss between $\mathcal{F}(I_t)$ and $\mathcal{F}(\hat{I}'_{t+i \rightarrow t})$ at deeper layers, the moving object segmentation is more close to the difference between the segmentation outputs of $I_t$ and $\hat{I}'_{t+i \rightarrow t}$ (effectively removing overlapping regions), resulting in poor performance. The threshold $\tau$ to decide whether an object is moving (as in Eq.~(\ref{eq:IsMoving})), is empirically determined. Surprisingly, the threshold $\tau$ has minimal impact on the performance. The mean IoU varies slightly, between $0.76-0.8$ for $\tau \in [0.1,0.9]$. For all the experiments, we set $\tau$ to $0.7$.
%
%
\setlength{\textfloatsep}{5pt plus 1.0pt minus 5.0pt}
\subsection{Inpainting}
Initially, we train the inpainting network proposed by Yu~\etal~\cite{Yu2018GenerativeII}. However, we do not use the refinement network as we do not observe any performance improvements. As input we supply 16 channels, 5 RGB images and a binary mask. However, no real ground truth is present for an input image, \ie we do not have an image taken at the exact same location without that moving object being there. Reprojected images could serve as ground truth, but have artifacts and a lower visual quality. Therefore, we randomly remove regions from the images (excluding regions of sky and recording vehicle) to generate ground truth. Instead of randomly selecting shapes for inpainting, the removed regions have the shapes of moving objects (obtained from moving object segmentation). The shapes of moving objects are randomly re-sized to different scales so as to learn to inpaint objects of different sizes.

\begin{figure*}[t!]
    \centering
    \begin{subfigure}{0.333\textwidth}
        \centering
        \includegraphics[width=164pt, height=140pt, trim= {0 10pt 0 14pt}, clip]{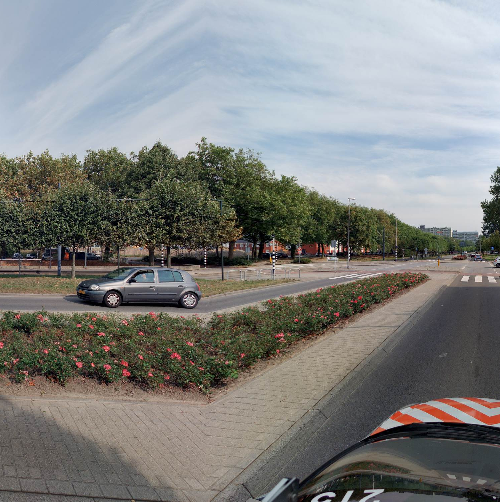}
    \end{subfigure}%
    \begin{subfigure}{0.333\textwidth}
        \centering
        \includegraphics[width=164pt, height=140pt, trim= {0 10pt 0 14pt}, clip]{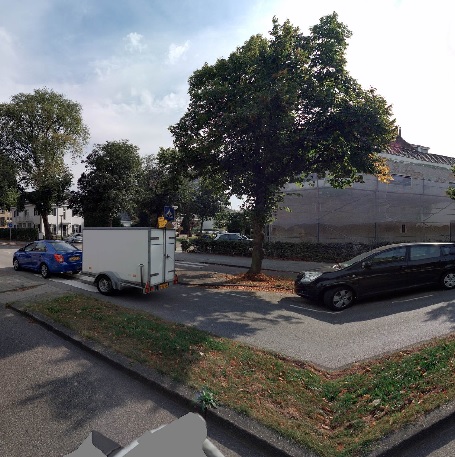}
    \end{subfigure}%
    \begin{subfigure}{0.333\textwidth}
        \centering
        \includegraphics[width=164pt, height=140pt, trim= {0 10pt 0 14pt}, clip]{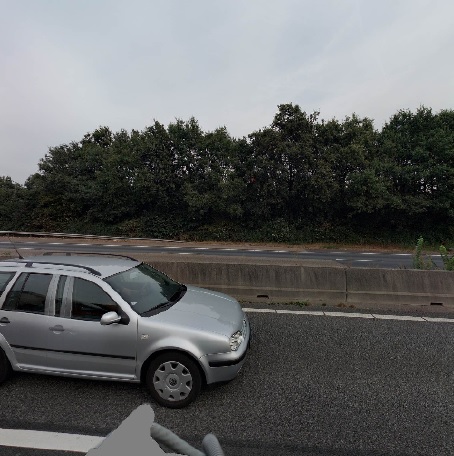}
    \end{subfigure}
    \begin{subfigure}{0.333\textwidth}
        \centering
        \includegraphics[width=164pt, height=140pt, trim= {0 10pt 0 14pt}, clip]{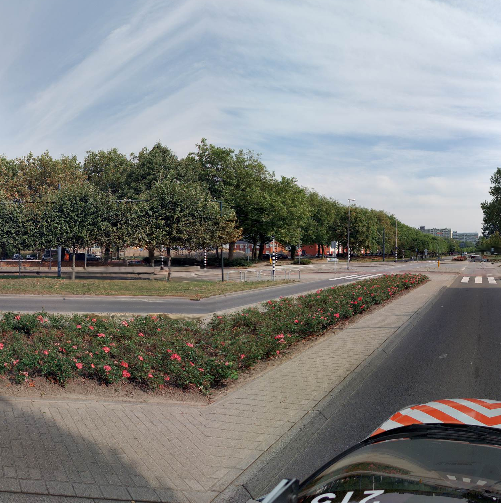}
    \end{subfigure}%
    \begin{subfigure}{0.333\textwidth}
        \centering
        \includegraphics[width=164pt, height=140pt, trim= {0 10pt 0 14pt}, clip]{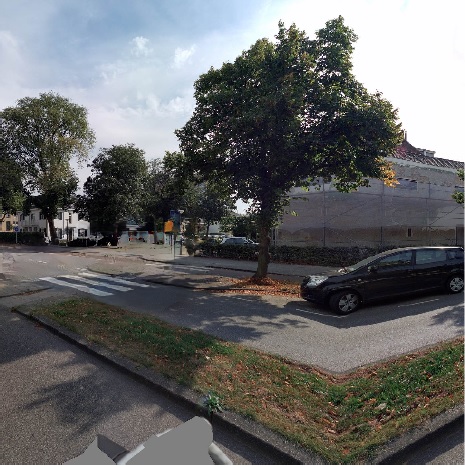}
    \end{subfigure}%
        \begin{subfigure}{0.333\textwidth}
        \centering
        \includegraphics[width=164pt, height=140pt, trim= {0 10pt 0 14pt}, clip]{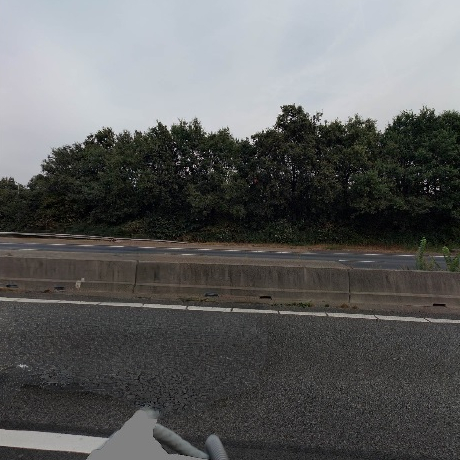}
    \end{subfigure}
    \caption{Inpainted results after removing objects obtained from moving object segmentation (bottom row). Input images are shown in the top row.
    Inpainting in the second and third column have a slight ghosting effect. Participants have an average confidence of 20\%, 62.9\% and 80\% (column 1-3) that it is an inpainted image.}
    \label{fig:Inpainting examples}
    \vspace{-2ex}
\end{figure*}

To provide an implicit attention for the inpainting GAN, instead of feeding in the complete reprojected images, we select only pixels from the non-empty regions of the binary mask from the other views. However, simply feeding in selected regions from other views have a drawback, since a moving object that is partially visible in other views, is also projected to the non-empty region. This is undesirable as it causes the inpainting network to learn unwanted moving objects from other views. Therefore, the moving objects from other views are removed prior to projecting pixels from other views. Therefore, the final input for the generator is [$I_t, B_t^h, \hat{I}^r_{t' \rightarrow t}$], where $[a, b]$ denotes the concatenation operation of $b$ after $a$. Optimization is performed using the Adam optimizer~\cite{kingma2014adam} with a batch size of~8 for both the discriminators and the generator. The learning rate is set to 10$^{-5}$ and is trained for 200~epochs. The discriminator-to-generator training ratio is set to~2. The inpainted results after moving object segmentation are shown in Fig.~\ref{fig:Inpainting examples}.
%
\vspace{-4mm}
\paragraph{Evaluation} Evaluation metrics such as Inception score (IS)~\cite{salimans2016improved}, MS-SSIM~\cite{snell2017learning} and Birthday Paradox Test~\cite{arora2018do} for evaluating GAN models are not suitable for inpainting as inpainting focuses on filling in background rather than its capacity to generate diverse samples. In the case of Fr\'echet Inception Distance (FID)~\cite{heusel2017gans} and IS~\cite{salimans2016improved}, a deep
network is trained such that it is invariant to image transformations and artifacts making it unsuitable for image inpainting task as these metrics have low sensitivities to distortions.

For evaluation, we use both PSNR and $\mathit{L_1}$ loss comparing the ground-truth image against the inpainted image on a test set of 1000 images. In our case, these metrics are suitable as they measure the reconstruction quality from other views rather the plausibility or diversity of the inpainted content. However, applying reconstruction losses as a evaluation metric favors multi-view based inpainting. As we use multi-view information for inpainting, it is obvious that the results can be improved significantly making it difficult for a fair comparison. Nevertheless, we report the results in PSNR and $\mathit{L_1}$ losses on the validation set. We obtain a PSNR and $\mathit{L_1}$ loss of \SI{27.2}{\decibel} and 2.5\% respectively. 

\begin{figure}[t!]
    \centering
    \includegraphics[width=0.97\linewidth]{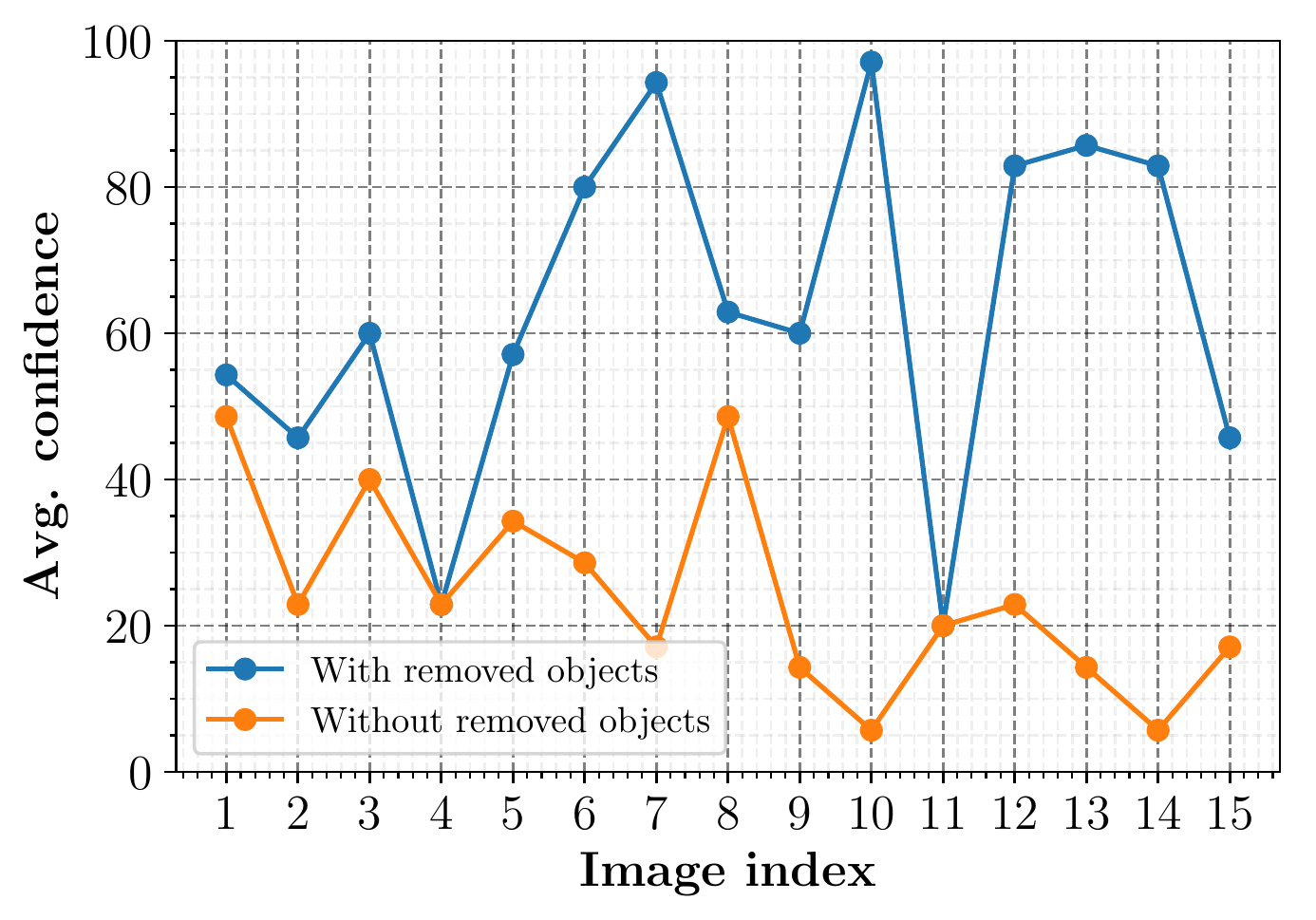}
    \caption{The average confidence per image of survey participants indicating if an image has an inpainted region. The blue line is for images that have objects removed and orange for unaltered images.}
    \label{fig:ResponseCollection}
\end{figure}

As a final experiment to assess overall quality, we have conducted a survey with 35~professionals within the domain. We have asked the participants to perform a strict quality check on 30~randomly sampled image tiles out of which 15 of them are inpainted after moving object detection. The number of tiles in which a moving object is removed is not revealed to the participants. Each participant is asked to observe an image tile for approximately 10~seconds and then determine if a moving object has been removed from the tile. Participants are also informed to pay close attention to misplaced shadows, blurry areas and other artifacts. The results of the survey is shown in Fig.~\ref{fig:ResponseCollection}. \CSe{In total of 1050 responses were collected from 35 participants, 333 (31.7\%) responses identified the true positives (inpainted images identified correctly as inpainted), 192 (18.3\%) as false negatives (inpainted images not recognized as inpainted), 398 (37.9\%) as true negatives (not inpainted and identified as not inpainted) and 127 (12.1\%) as false positives (not inpainted but recognized as inpainted).} Note that combination of true positives and false negatives are disjoint from the combination true negatives and false positives. The participants have an average confidence of 63.4\%~$\pm$~23.8\% that a moving object(s) was inpainted in the images where objects were removed (average of responses in blue line of Fig.~\ref{fig:ResponseCollection}). However, it is interesting to note that in cases when no object is removed, they have a confidence of 24.2\%~$\pm$~13.5\% stating that an object(s) is removed and inpainted (average of responses in orange line of Fig.~\ref{fig:ResponseCollection}).

\CSe{Clearly, in most cases with meticulous observation, participants are able to discern if an object is removed.} However, we also observe high deviation in the responses of images where objects are removed and hence we inspect images that have poor scores (high confidence from participants stating that it is inpainted). Few of the worst performing results (confidence higher than 90\% on the blue line of Fig.~\ref{fig:ResponseCollection}) are shown in Fig.~\ref{fig:FailureCases}. We have found that the worst results (Image~7, Image~10 with average confidence of 94.3\%, 97.1\%) have the strongest artifacts. However, in the other cases (average confidence of 82.9\%, 85.7\% and 82.9\%), we note that minuscule errors such as slight variations in edges, lighting conditions, shadows, etc. are the reasons why participants are able to distinguish the inpainted examples. Although the poor cases are reported with high confidence, we believe that such artifacts would be hardly noticed in reality if not explicitly searched. Despite these cases, this framework ensures complete privacy, alleviates blurring artifacts and removes occluded regions which is beneficial for commercial purposes.
\section{Discussion}
\label{sec:Discussion}
\begin{figure}[!t]
    \centering
    \begin{subfigure}{0.24\textwidth}
        \centering
        \includegraphics[width=118pt, height=77pt, trim= {0 19pt 0 19pt}, clip]{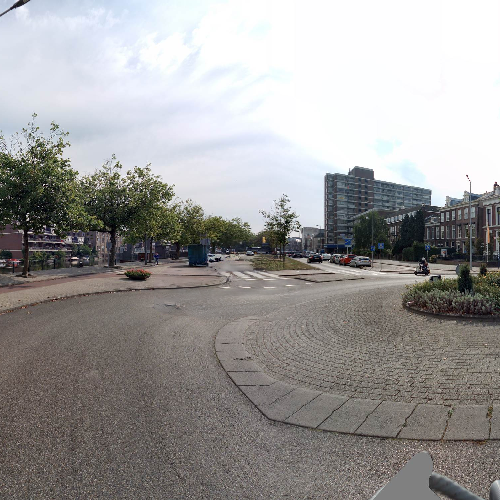}
    \end{subfigure}%
    \begin{subfigure}{0.24\textwidth}
        \centering
        \includegraphics[width=118pt, height=77pt, trim= {0 19pt 0 19pt}, clip]{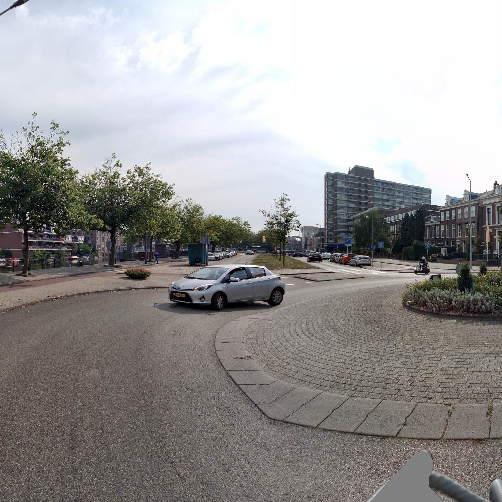}
    \end{subfigure} 
    \begin{subfigure}{0.24\textwidth}
        \centering
        \includegraphics[width=118pt, height=77pt, trim= {0 19pt 0 19pt}, clip]{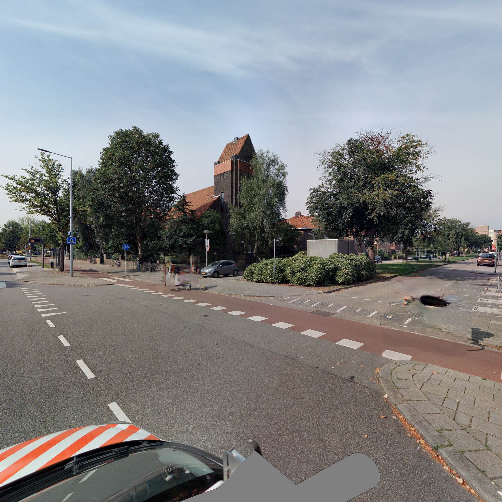}
    \end{subfigure}%
    \begin{subfigure}{0.24\textwidth}
        \centering
        \includegraphics[width=118pt, height=77pt, trim= {0 19pt 0 19pt}, clip]{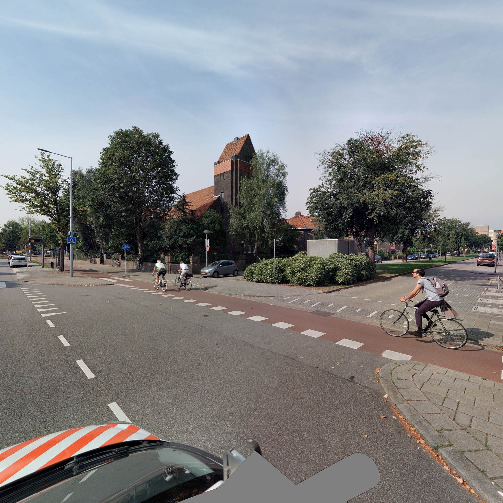}
    \end{subfigure}
    \begin{subfigure}{0.24\textwidth}
        \centering
        \includegraphics[width=118pt, height=77pt, trim= {0 19pt 0 19pt}, clip]{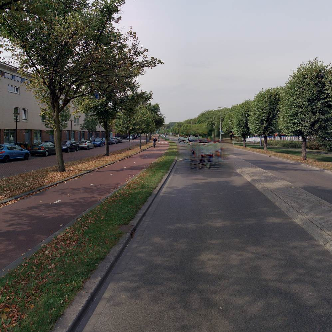}
    \end{subfigure}%
    \begin{subfigure}{0.24\textwidth}
        \centering
        \includegraphics[width=118pt, height=78pt, trim= {0 19pt 0 19pt}, clip]{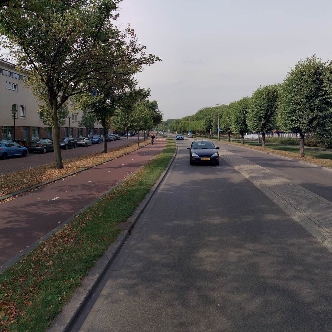}
    \end{subfigure}
    \vspace{-1ex}
    \caption{Worst performing results from the survey (high confidence that object is removed). Rows 1-3 with average confidence of 82.9\%, 94.3\% and 97.1\% respectively. }
    \label{fig:FailureCases}
\end{figure}

Although the proposed framework is a good alternative to blurring, it is by no means perfect. The moving object segmentation algorithm invokes a few challenges. Since the moving object detection is class agnostic, there are false positives from certain objects such as traffic signs and light poles. The comparison between features from different viewpoints of a traffic sign (front and back) leads to false positives. Similarly, poles might be detected as moving since small camera position errors lead to a minor mismatch of depth pixels during reprojection. However, we are able to suppress these false positives by combining the outputs from FC-VGGNet. Even for a wide range of $\tau$ values [0.1 - 0.9], mIoU varies only by 4\% ensuring the reliability and robustness of the method. The proposed method may fail when there is an overlap of moving and static objects. For example, a car driving in front of parked vehicles can results in all objects classified as moving or non-moving. However, this can be mitigated by applying instance segmentation.
\vspace{-4mm}
\paragraph{Limitations} Poor results occur in a few cases when a driving vehicle is in the same lane as the recording vehicle (Fig.~\ref{fig:FailureCases}, row 3). The moving object completely occludes all the views making it difficult for multi-view inpainting. In such a scenario inpainting based on context would be an alternative, however, this does not guarantee a genuine completion of the image. In non-commercial application, this is still a viable solution. Even though few inpainting artifacts such as shadows, slightly displaced edges are visible, we argue that they are still a better alternative to blurring as it ensures complete privacy and far less noticeable artifacts (Fig.~\ref{fig:Inpainting examples}, column 3 and Fig.~\ref{fig:FailureCases}, row 1). \CSe{As the method does not explicitly target shadow, this too may reveal privacy-sensitive information in rare cases.}

\vspace{-1mm}
\section{Conclusion}
\vspace{-1mm}
We \DMG{presented} a framework that is \DMG{an} alternative for blurring in the context of privacy protection \DMG{in street-view images}. The proposed framework comprises of novel convolutional feature based moving object detection algorithm that is coupled with a multi-view inpainting GAN to detect, remove and inpaint moving objects. We \DMG{demonstrated} through the multi-view inpainting GAN that legitimate information of the removed regions can be learned which is challenging for a standard context based inpainting GAN. 
\DMG{We also evaluated overall quality by means of a user questionnaire}. Despite the discussed challenges, the inpainting results of the proposed method are often hard to notice and ensures complete privacy. Moreover, the proposed approach mitigates blurring artifacts and removes occluded regions which are beneficial for commercial applications. Although most of the current solutions rely on blurring, we believe that the future of privacy protection lies in the direction of the proposed framework.

{\small
\bibliographystyle{ieeetr}
\bibliography{egbib}
}

\end{document}